%% file: main.tex
\documentclass[a4paper]{article}

\usepackage{INTERSPEECH_v2}
\usepackage{tipa}
\usepackage{color}
\usepackage{tabularx}
\usepackage{cite}
\usepackage{arydshln}
\usepackage{url}

\title{Automatic Measurement of Pre-aspiration}
\name{Yaniv Sheena$^1$, M\'i\v{s}a Hejn\'a$^2$, Yossi Adi$^1$, Joseph Keshet$^1$}
\address{$^1$Department of Computer Science, Bar-Ilan University, Ramat-Gan, Israel\\
$^2$Department of English, Aarhus University, Denmark}
\email{yaniv.sheena@live.biu.ac.il, misa.hejna@cc.au.dk, jkeshet@cs.biu.ac.il}

\newcommand{\half}{\frac{1}{2}}
\newcommand{\X}{\mathcal{X}}
\newcommand{\x}{\mathbf{x}}
\newcommand{\sx}{\bar{\x}}
\newcommand{\ts}{t_s}
\newcommand{\tsi}{t_{s_i}}
\newcommand{\pts}{\hat{t}_s}
\newcommand{\te}{t_e}
\newcommand{\tei}{t_{e_i}}
\newcommand{\pte}{\hat{t}_e}

\newcommand{\R}{\mathbb{R}}

\newcommand{\T}{\mathcal{T}}

\newcommand{\w}{\mathbf{w}}
\newcommand{\bphi}{\boldsymbol{\phi}}

\renewcommand{\eqref}[1]{Eq.~(\ref{#1})}

\newcommand{\D}{\Delta}

\DeclareMathOperator*{\argmin}{arg\,min}
\DeclareMathOperator*{\argmax}{arg\,max}

\begin{document}

\maketitle
\begin{abstract}
Pre-aspiration is defined as the period of glottal friction occurring in sequences of vocalic/consonantal sonorants and phonetically voiceless obstruents. We propose two machine learning methods for automatic measurement of pre-aspiration duration: a feedforward neural network, which works at the frame level; and a structured prediction model, which relies on manually designed feature functions, and works at the segment level. The input for both algorithms is a speech signal of an arbitrary length containing a single obstruent, and the output is a pair of times which constitutes the pre-aspiration boundaries. We train both models on a set of manually annotated examples. Results suggest that the structured model is superior to the frame-based model as it yields higher accuracy in predicting the boundaries and generalizes to new speakers and new languages. Finally, we demonstrate the applicability of our structured prediction algorithm by replicating linguistic analysis of pre-aspiration in Aberystwyth English with high correlation.
\end{abstract}
\noindent\textbf{Index Terms}: pre-aspiration, feedforward neural network, structured prediction, laboratory phonology

\label{intro}
\input{01_intro}

\label{problem_set}
\input{02_problem_set}

\label{model}
\input{03_learning}

\label{dataset}
\input{04_data_set}

\label{experiments}
\input{05_experiments}

\label{discussion}
\input{06_discussion}

\bibliographystyle{IEEEtran}

\bibliography{mybib}

\end{document}

%% file: 01_intro.tex

\section{Introduction}
Pre-aspiration is a period of (primarily) glottal friction, which is found in the sequences of sonorants and phonetically voiceless obstruents prior to the release of the consonant, e.g., in Welsh English \textit{kit} /k\textsuperscript{h}\textipa{I}\textsuperscript{h}t\textsuperscript{s}/, \textit{miss} /m\textipa{I}\textsuperscript{h}s/, \textit{milk} /m\textipa{I}\textipa{\r*l}k\textsuperscript{h}/, etc. Although pre-aspiration was previously considered rare \cite{Hel2002, Sil2003}, with the advances in recording technology it has been recently reported to occur in increasingly more languages and varieties thereof \cite{Clay2015, dCan2012, DF1999, GS2010, H2015, HSc2015, JL2003, Ket2015 ,R1998, S2010, SH2004, W2007, Dwyer2000}. Recent studies of pre-aspiration have revealed that the phenomenon is very individual \cite{H2015,Hel2002,RD2013}, sensitive to sex/gender \cite{NSS2013}, and can be affected by age \cite{DF1999,H2015} and the first language of the speaker \cite{NSS2013}. 

Most of the work on pre-aspiration has been based on subjective, labor-intensive manual annotation. The vast majority of researchers have coded pre-aspiration manually by identifying its boundaries, usually based on a manual segmentation of the signal into segmental and subsegmental intervals \cite{Hel2002,dCan2012,H2015,JL2003,Ket2015,S2010,SH2004,NSS2013,H2016a,RH2008,Hel2004,vDomm1998}. 

In this paper we tackle the problem of automatic measurement of pre-aspiration duration. We propose two algorithms for this purpose. The first one is a feedforward neural network, which works at the frame-level, and the second algorithm is based on structured prediction techniques which rely on highly tuned feature maps which were specifically designed for the task, and which works at the segment level.

As far as we know this is the first attempt to automatically detect pre-aspiration. This work is based on our ongoing work on designing and developing machine learning algorithms for automatically measuring with high precision, phonetic properties of speech at the level of human inter-transcriber reliability \cite{sonderegger2012automatic,adi2016automatic,adi2016prevoicing,dissen2016formants,adi2017segmentor}. Our methods rely on several advances over existing computational systems: novel representations of the speech signal and new structured prediction and deep learning algorithms. These automatic methods allow for low-cost replication of phonetic studies and expand the range of empirical and theoretical issues we can address.

The code is available to used by the research community. It can be downloaded from \url{https://github.com/MLSpeech/AutoPreaspiration}.

%% file: 02_problem_set.tex

\section{Problem Setting}

In the context of a typical phonological study, given a portion of an acoustic signal of an arbitrary length, the goal of an automatic pre-aspiration measurement algorithm is to predict the onset and offset times of pre-aspiration as accurate as possible. In this work we assume that the acoustic signal contains exactly one coded obstruent within a word and the signal starts before the expected pre-aspiration, i.e., during the previous phoneme. 

We turn to describing the problem formally. Throughout the paper, scalars are denoted using lower case Latin letters, e.g., $x$, and vectors using bold face letters, e.g., $\x$. A sequence of elements is denoted with a bar $\sx$ and its length is written as $|\sx|$. 

The acoustic input is represented as a sequence of feature vectors, $\sx = (\x_1,\ldots,\x_T)$, where each vector, $\x_t$, is $D$-dimensional vector which represents the acoustic content of the $t$-frame $(1\leq t \leq T)$. The domain of the feature vectors is denoted as $\X\subset\R^D$. The acoustic input is of an arbitrary length, hence the number of frames, $T$, is not fixed. We denote by $\X^*$ the set of all finite-length sequences over $\X$. 

\begin{figure}  
   \includegraphics[width=\linewidth]{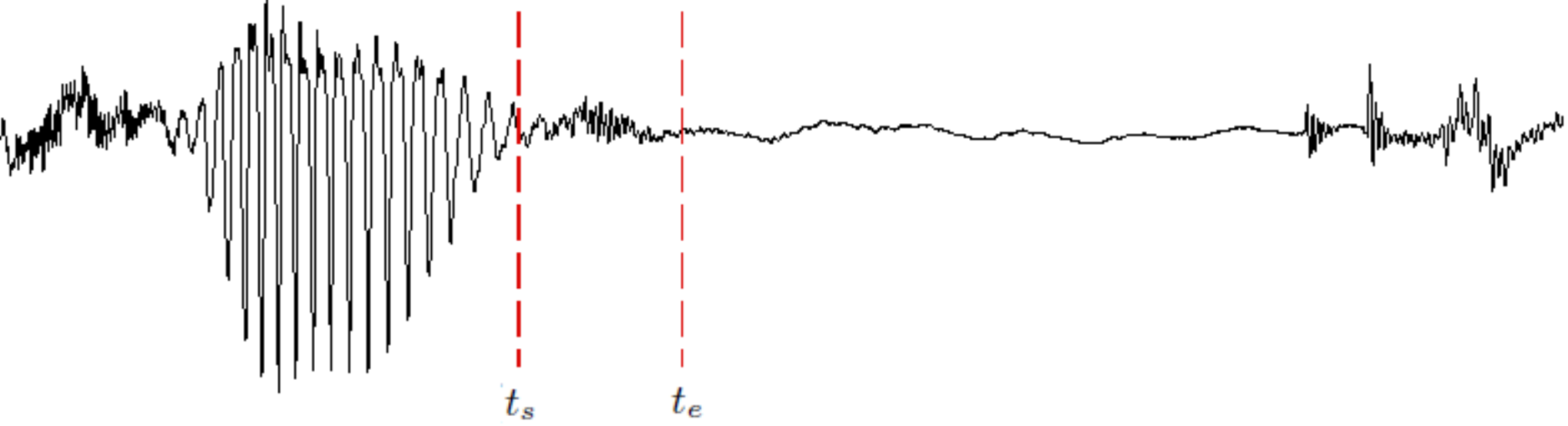}
   \footnotesize{\caption{\label{fig:examp}\it Annotation example of pre-aspiration. The signal consists of a vowel followed by a plosive of the word ``cook''.}}
\end{figure}

Each acoustic input is associated with a timing pair: the \emph{onset} of the pre-aspiration, $t_{s} \in \T$, and the \emph{offset} of the pre-aspiration, $t_{e} \in \T$, where $\T = \{{1}, \ldots, T\}$. To sum up, given the speech interval $\sx$, our goal is to learn a function $f$ from the domain of all acoustic inputs $\X^*$ to the domain of all possible onset offset pairs $\T^2$. Our notation is depicted in Figure~\ref{fig:examp}.

%% file: 03_learning.tex

\section{Learning Apparatus}

In this section, we describe how we learn the prediction function from a training set of examples. We denote the training set of $m$ examples by $S=\{(\sx^i,\ts^i,\te^i)\}_{i=1}^{m}$, where the $i$-th example is composed of a sequence of acoustic features $\sx^i$ labeled with an onset and offset pair, $(\ts^i, \te^i)$.

Let us denote the predicted onset and offset pair by $(\pts, \pte)$, namely $(\pts,\pte) = f_{\theta}(\sx)$, where $\theta$ denotes the set of parameters of the prediction function $f$. In order to assess the quality of the prediction we use a \emph{task loss} function, $\gamma\left( (\te,\ts),(\pte,\pts) \right)$, which returns a real positive number that measures how the prediction pair $(\pts,\pte)$ is close to the manually annotated pair $(\te,\ts)$. Our goal in learning is to find the set of parameters $\theta$ so as to minimize the expected task loss.

We start by describing the acoustic features, and then two different machine learning algorithms that learn the parameters of function $f$. The first algorithm works at the frame level and hence cannot aim at minimizing a global task loss function, while the second algorithm is aimed directly at minimizing the expected task loss. These differences are reflected in the accuracy level of the different algorithms.


\subsection{Acoustic features}

For both machine learning models, we extract the same set of features. Consider the acoustic input $\sx = (\x_1,\ldots,\x_T)$ consisting of $T$ frames, where each acoustic feature vector $\x_t$ consists of $D$ features. Similar to \cite{sonderegger2012automatic}, we extract eight ($D$=8) acoustic features from the speech signal every 1 ms. The first four features refer to the total spectral energy ($E_{\textrm{total}}$), energy between 50-1000 Hz ($E_{\textrm{low}}$), energy above 3000 Hz ($E_{\textrm{high}}$), and Wiener entropy ($H_{\textrm{wiener}}$) --- all are based on the short-time Fourier transform (STFT) taken every 1 ms with a 5 ms Hamming window. The fifth feature, $P_{\textrm{max}}$, is the maximum of the power spectrum calculated in a region from 6 ms before to 18 ms after the frame center. The sixth feature ($R_l$) is the pitch of the signal using a real-time pitch detector \cite{sha2004real}. The seventh feature is the $0/1$ output of a voicing detector based on the RAPT pitch tracker \cite{talkin1995robust}, smoothed with a 5 ms Hamming window ($V$). The last feature is the number of zero crossings ($ZC$) in a 5 ms window around the frame center. 

\subsection{Frame-based model}
\label{sub_frame_based}

Our first model works at the frame level. We train a binary classifier such that given an input speech frame $\x_t$ predicts whether or not it is associated with a pre-aspiration event. In order to take advantage of the local temporal context of each time-frame, the input of this binary classifier is based on five concatenated feature vectors $(\x_{t-2}, \x_{t-1}, \x_{t}, \x_{t+1}, \x_{t+2})$ rather than a single frame. We use a feedforward neural network. We tried several network architectures, and chose the one that performed best on a validation set. The network consists of an input layer of 40 units (recall that we extract $D=8$ feature per frame, and have 5 consecutive frames), a hidden layer of 40 units with ReLU activation function and a dropout rate of 0.3, and one output unit followed by a sigmoid. The network was trained to minimize the binary cross entropy loss using the gradient descent optimization algorithm. 

At inference time, we use the trained model sequentially over the input frames to produce a sequence of predictions. Since this sequence can be noisy, we smooth it out using a moving average, followed by a binary mapping that uses a threshold for each such average. In the final step, we search the longest subsequence of frames that were predicted as associated with pre-aspiration, and output its boundaries as the onset and offset of the pre-aspiration.

\subsection{Structured model}

The second algorithm takes advantage of the input as a whole segment, hence we can introduce feature maps such as typical pre-aspiration duration, mean energy during the presumed pre-aspiration compared to the mean energy before or after the pre-aspiration. Similar to previous work in structured prediction
\cite{taskar2003max,tsochantaridis2004support}, the function $f$ is constructed from a predefined set of $N$ feature maps $\{\phi_i\}_{i=1}^N$, each of the form $\phi_i:\X\times\T\times\T\times\to\R^{N}$, and a weight vector $\w\in\R^N$.  The function is a linear predictor of the following form
\begin{equation}
(\pts,\pte) = \argmax_{(\ts,\te)} \w\cdot\bphi(\sx,\ts,\te),
\end{equation}
where we have used vector notation for the feature maps
$\bphi=(\phi_1,\ldots,\phi_N)^\top$. This vector-valued function is used to map the variable length of input speech along with a presumed onset-offset pair to an abstract vector space in $\R^N$ (described in \ref{sub_feature_map}). 

The algorithm presented here aims to find the weights $\w$ that minimize the expected task loss function $\gamma$ which measures the distance between predicted and manually coded labels. In a similar manner to \cite{sonderegger2012automatic}, we define the task loss function as follows:
\begin{equation}\label{eq:cost}
\gamma\left( (\te,\ts),(\pte,\pts) \right) 
= \max\{| (\pte-\pts) - (\te-\ts)|-\epsilon,0\}.
\end{equation}
That is, only the differences between the predicted pre-aspiration and the  manually labeled pre-aspiration that are greater than a threshold $\epsilon$ (in milliseconds), are penalized.  This task loss function takes into account that manual measurements are not usually exact, and $\epsilon$ can be adjusted according to the level of measurement uncertainty in a dataset.

The weight vector $\w$ is learned using an iterative algorithm based on the \emph{Passive-Aggressive} family of algorithms for structured prediction\cite{crammer2006online-short}. Let $\w_t$ be the weight vector after the $t$-th iteration, and let $\w_0 = \mathbf{0}$.  At each iteration a single example $(\sx^i,\ts^i,\te^i)$ is considered, and the current weight vector $\w_t$ is updated by finding the solution to the following optimization problem:
\begin{multline}\label{eq:PA_I} 
\w_{t+1}=\argmin_{\w,\xi \geq 0} ~~ \half \Vert\w - \w_{t}\Vert^2 + C\xi \\ 
\textrm{s.t.} ~~\w \cdot  \bphi(\sx^i,\ts^i,\te^i) - 
\w \cdot \bphi(\sx^i,\tilde{\ts},\tilde{\te})  \geq \gamma_i - \xi ~,
\end{multline}
where $\gamma^i=\gamma\left( (\ts^i,\te^i),(\tilde{\ts},\tilde{\te})
\right)$, $C$ serves as a trade-off parameter between loss and regularization  minimization, and $\xi$ is a non-negative slack variable. 

The optimization problem tries to keep the new weight vector $\w$ close to the previous weight vector $\w_{t}$ while satisfying the constraint that the score of the manually annotated onset-offset pair $\w \cdot  \bphi(\sx^i,\ts^i,\te^i)$ will be higher than a different set of onset-offset, $\w \cdot \bphi(\sx^i,\tilde{\ts},\tilde{\te})$, where the onset-offset pair $(\tilde{\ts},\tilde{\te})$ in the constraint is the most violated pair \cite{tsochantaridis2004support}, namely:
\begin{equation}
(\tilde{\ts},\tilde{\te}) = \argmax_{(\tilde{\ts},\tilde{\te})} \w\cdot\bphi(\sx^i,\tilde{\ts},\tilde{\te}) + \gamma^i.
\end{equation}

\begin{table*}[t]
	\centering
	\caption{Summary of the feature maps. The rows represent the type of calculations that should be performed. The columns are the eight acoustic  features. $F$ in row $i$ and column $j$ indicates that there is a feature map of
  type $i$ for feature $x_j$; $\D$ indicates that there are three feature
  maps of type $i$ for the local difference of feature $x_j$,
  evaluated at $s=5, 10, 15$ \cite{sonderegger2012automatic}. For example, the $F$ in the last row, column $E_{\textrm{high}}$ 
  denotes a feature-map that gets the mean of the value of $E_{\textrm{high}}$ over 50 frames starting from $\te$.}  
  	\label{tab:Feature-Map}
    \renewcommand{\arraystretch}{0.7}
    \vspace{-5pt}
  	\begin{tabularx}{\textwidth}{lcccccccc}
  	\toprule
   	Feature map  type & $E_{\textrm{total}}$ & $E_{\textrm{low}}$ & $E_{\textrm{high}}$ &  $H_{\textrm{wiener}}$
   	& $P_{\textrm{max}}$ & $R_l$ & $V$ & $ZC$ \\
   	\midrule
   	Value at $\ts$ &$F,\D$& $\D$ & $F,\D$ & $F,\D$ &$\D$& $F,\D$ & $\D$ & $\D$ \\
   	Value at $\te$ & $F,\D$ & $F,\D$ &$F,\D$& $F,\D$ & $F,\D$ & $F,\D$ && \\
   	Mean \& max over $(\ts,\te)$ & &&&&$F,\D_{5},\D_{10}$&& \\
   	Mean over $(\ts,\te)$ - mean over $(\te,\te+50)$ & && $F$ &$F$& & & & $F$ \\
   	Mean over $(\te,\te+50)$ &&&$F$&$F$&& $F$ & $F$ & \\
   	Max over $(\te,\te+50)$ &&&$F$&$F$&&& \\
   	\hline
	\end{tabularx}
\end{table*}

\subsection{Feature maps for the structured model} \label{sub_feature_map}

The feature maps are built from local differences, cumulative mean and max over subsets features with respect to the pre-aspiration boundaries. They were chosen based on manual inspection of the values and the trends of the features in intervals near $\ts$ and $\te$, and they reflect the knowledge about the  problem of pre-aspiration measurement, which in our case is based on \cite{H2015}.  

As an example for such a feature map we considered the observation that when pre-aspiration is immediately followed by a silent interval in the context of plosives, we expect that high frequencies (above 3000 Hz) will decrease compared to the interval of pre-aspiration, which has high energy presence in these frequencies. In order to express this observation, we define feature maps that compute the differences of the means of $E_{\textrm{high}}$ and $H_{\textrm{wiener}}$ over a post pre-aspiration interval $(\te,\te + 50)$ and the pre-aspiration interval $(\ts,\te)$.

Another key observation is that $\ts$ usually comes right after a formant structure which becomes less distinct and usually indicates that voicing is ending. Hence a set of feature maps are based on the local differences of 5, 10 and 15 milliseconds over the acoustic features $P_{\textrm{max}}$ and $V$ in order to allow our algorithm to capture these changes. A full specification of $\bphi$ is shown in Table~\ref{tab:Feature-Map}.

%% file: 04_data_set.tex

\section{Data Set}

The data is composed of 5,297 examples of 16 speakers of English from the town of Aberystwyth within mid Wales. All of the speakers were born and raised in the town and are L1 Welsh speakers. The speakers presented are 10 females and 6 males, and their age range spans from 22--91 to allow for preliminary generational comparisons. The data itself consisted of a list of words that the participants read; each word contained a sequence of a vowel and a post-tonic plosive. Each word was read once in isolation and twice in a frame sentence. For more details on the segmental and prosodic characteristics of the tokens related to aspects such as vowel height, place and manner of articulation of the plosive, stress, and foot-position see \cite{H2015}. The recordings were obtained with an H4 Zoom Handy Recorder (sampled at 44.1kHz) in conjunction with the head-mounted AKG C520 Microphone, which was attached to the speaker's head and ensured a constant distance from the mouth, irrespective of the speaker's movements.

%% file: 05_experiments.tex

\section{Experiments}

We evaluated the performance of our algorithms using three main methods. First, we compared the predictions of the frame-based algorithm and of the structured algorithm to the manual annotations. Then we tested how the structured prediction algorithm, which was superior to the frame-based algorithm, generalizes on new set of speakers and languages. Finally, we replicated a linguistic study on the model predictions and compared it to the results obtained from the manually annotated data. All of the experiments in this section are based on models which were trained using search windows of 50 ms before the labeled left boundary ($\ts$) and 60 ms after the right boundary ($\te$), where we tried several window-sizes and the effects were insignificant.

\subsection{Models performance}
\label{Sub_performance}

We evaluated both algorithms on the corpus described in the previous section using 5-fold cross-validation. For each fold, 15\% of the data served as a validation set. We regularized both algorithms using early stopping. For the frame-based algorithm, we balanced the data by randomly dropping negative examples, as the amount of such examples was significantly greater than the amount of positive examples. For the structured model we used $C=50$, which was chosen on a validation set, and $\epsilon$ in the task loss was set to 2 ms.

We compared the models by reporting the percentage of examples in the test set with automatic/manual differences which were less than a time tolerance, where we used tolerances of 5, 10, 15, and 20 ms. We also report the average error in the prediction of the onset and the offset. The results are given in Table~\ref{tab:Prop_models}. Results should read as follows: The percentage of correctly predicted pre-aspirations within a threshold of 2 ms was 43.3\% for the frame-based algorithm and 56.2\% for the structured algorithm. The mean difference between the predicted and manually-annotated onset was 4.4 ms for the frame-based algorithm and 2.6 ms for the structured algorithm. 

\begin{table}[h]
\small
\renewcommand{\arraystretch}{1.4}
	\centering
	\caption{\it Results for the two algorithms. The first 4 columns are the percentage of accurately predicted pre-aspirations within a given threshold (thresholds in ms, values are percentages). The remaining columns are the mean distance between the boundaries of manual/automatic pre-aspirations in ms. }
  \vspace{-5pt}
  \label{tab:Prop_models}
  \begin{tabular}{lcccc|cc}
    \hline
	Algorithm & $\le$5 & $\le$10 & $\le$15
    & $\le$20 & $\D\ts$ & $\D\te$ \\
    \hline
	Frame-based  & 43.3 & 74.3 & 88.9 & 95.3 & 4.4 &  5.3 \\
    Structured & 56.2 & 84.8 & 93.1 & 96.3 & 2.6 &  4.9 \\ 
    \hline
  \end{tabular}
\end{table}

In Table~\ref{tab:Mean_std_models} we report the mean and the standard deviation of the 
manually annotated pre-aspiration duration and the predicted pre-aspiration duration for both algorithms. 

\begin{table}[h]
\renewcommand{\arraystretch}{1.4}
	\small
	\centering
	\caption{\it Mean and standard deviation of pre-aspiration duration. Manually coded versus models' predictions.}
  \vspace{-5pt}
  \label{tab:Mean_std_models}
  \begin{tabular}{lcc}
    \hline
	Type & Mean (ms)&  Standard deviation (ms)\\
    \hline
    Manual      & 37.6  & 20.8  \\
    Structured  & 37.2 & 19.9  \\ 
	Frame-based & 42.0 & 17.8  \\ 
    \hline
  \end{tabular}
\end{table}

Results suggest that the structured algorithm is superior to the framed-based algorithm. This demonstrates the power of the structured algorithm, which uses a set of dedicated feature-maps each of which measures properties and trends of intervals surrounding the boundaries of the pre-aspiration, contrary to the frame-based method that is based on local windows. Henceforth, we conducted the rest of the experiments using only the structured algorithm.

\subsection{Generalization}

In order to test our algorithm generalization on data produced by different sources, we conducted several experiments. First, we performed leave-one-speaker-out (LOSO) using the 16 speakers of Aberystwyth English, i.e., the utterances of each speaker were tested individually with a structured model trained on the rest of the speakers. The averaged results are summarized in the first row of Table~\ref{tab:Lang_results}. Note that the results are very similar to the results shown in the second row of Table~\ref{tab:Prop_models}, where the data was randomly partitioned using 5-fold cross-validation.

Next, we evaluated our structured algorithm on a data set consisting of 607 speech intervals of the Welsh language (16 speakers from Bethesda, North Wales). The performance of the algorithm on the Welsh corpus using 4-fold cross-validation results are summarized in the second row of Table~\ref{tab:Lang_results}. Our goal was to compare the performance of the structured algorithm that was trained on one language on a different language. The last two rows in Table~\ref{tab:Lang_results} outline the performance of the algorithm that trained on Aberystwyth English but was tested on Welsh and vice versa.

It is interesting to note that there are different performance drops when languages are mismatched. When training on the Aberystwyth data and testing on Welsh data results drop by 7.2\% with a tolerance level of 10 ms. However, when tested on Aberystwyth data and training on the Welsh data we suffered only a 4.3\% decrease in results, again with a 10 ms tolerance level.

\begin{table}[h]
\renewcommand{\arraystretch}{1.4}
	\small
	\centering
	\caption{\it Mismatched training and test sets. LOSO stands for leave-one-speaker-out. Thresholds and time differences in ms, values are percentages.}
  \vspace{-5pt}
  \label{tab:Lang_results}
  \begin{tabular}{lcccc|cc}
    \hline
	Train/Test & $\le$5 & $\le$10  & $\le$15 & $\le$20 &  $\D\ts$ & $\D\te$ \\
    \hline
    Aber./Aber. LOSO\hspace{-1em} & 54.0 & 81.9 & 90.5 & 94.3 & 2.9 & 5.6\\
		\hdashline
    Welsh/Welsh & 52.7 & 76.7 & 87.1 & 91.7 & 5.3 & 4.5 \\
    Aber./Welsh  & 45.8 & 69.5 & 84.4 & 88.8 & 6.6 & 6.5 \\
    Welsh/Aber.  & 54.6 & 80.5 & 90.5 & 94.7 & 3.2 & 5.2 \\	 
    \hline
  \end{tabular}
\end{table}

\subsection{Linguistic analysis}

As the purpose of the algorithm is to reliably measure pre-aspiration duration for linguistic and language-related analyses, we also carried out an analysis of what language-internal and language-external factors affect pre-aspiration duration measured manually as opposed to pre-aspiration duration measured automatically with the algorithm. The outputs of four Linear Mixed Effects Models were compared. These models differed only in the dependent variable, which was (i) manually coded raw pre-aspiration duration, (ii) automatically coded raw pre-aspiration duration, (iii) manually coded normalized pre-aspiration duration, and (iv) automatically coded normalized pre-aspiration duration. The normalization method expressed the raw measurements as a percentage of the overall word duration.

The models shared the following independent variables: Age (continuous variable), Sex (with two levels: \emph{female} and \emph{male}), Vowel type (with eight levels: /a/ contrasted with /e/, /\textipa{I}/, /\textipa{6}, /\textipa{U}/, /\textipa{2}/, /\textipa{A:}/, and /o\textipa{:}/), Place of articulation of the plosive (with three levels: /p/, /t/, /k/), and Word position (with two levels: \emph{word-medial} and \emph{word-final}); Speaker and Word were selected as random effects. Forward difference coding was applied to the place of articulation. 

\begin{figure}  
   \includegraphics[width=\linewidth]{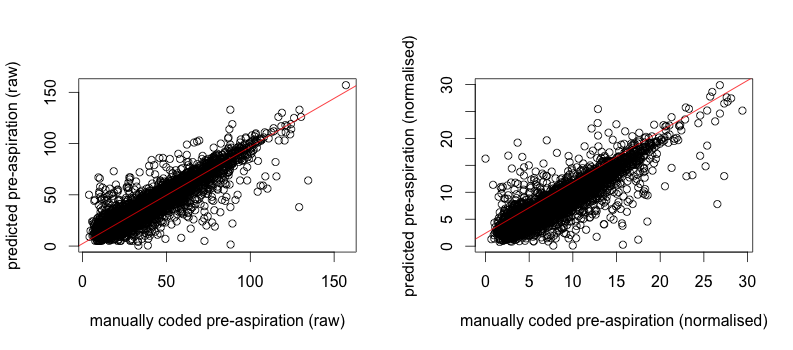}
   \footnotesize{\caption{\label{fig:correlations}\it Correlations between the manually coded and the predicted durations of pre-aspiration; raw (left) and normalized (right) values.}}
\end{figure}

The tests show the same results regarding language-internal variables, irrespective of whether normalized or raw data are used. In all cases, pre-aspiration intervals are longest with /a/ as opposed to the other vowels (p $<$ 0.0001), with the exception of /\textipa{6}/, which does not differ from /a/ in the durations of pre-aspiration with which it is associated. Furthermore, the duration of pre-aspiration increases as we move further back in the oral cavity (/p/ $<$ /t/ $<$ /k/; p $<$ 0.0001). In addition, pre-aspiration is longer foot-finally than medially (p $<$ 0.0001). The models also yield the same results concerning the external variable of sex in that no effect is found (p = 0.06-0.39). The only difference across the models is related to the variable of age, which comes out as significant when the dependent variable is manually coded normalized pre-aspiration ($<$ 0.05; as opposed to p = 0.06-1). Moreover, there is a very strong positive correlation between the predicted and the manually coded values (raw measurements: r = 0.87; normalized measured: r = 0.89; p $<$ 0.0001; Pearson's product moment correlation) as shown in Figure~\ref{fig:correlations}.

%% file: 06_discussion.tex
\section{Discussion}
We have presented a new trainable algorithm for automatic measurement of pre-aspiration. To our knowledge, this is the first attempt to develop an automatic tool for measuring pre-aspiration. We have shown that the structured algorithm outperforms the frame-based neural network algorithm. Furthermore, we reproduced the results of a linguistic analysis based on pre-aspiration, solely using our structured algorithm's predictions. Future work can involve extending the use of the algorithm to fricative context, automatically detecting whether there is pre-aspiration in a given interval, and applying the presented methods to lower quality data.

\section{Acknowledgements}
We wish to thank Jon Morris for letting us use his Welsh data.

%% file: main.bbl
\begin{thebibliography}{10}
\providecommand{\url}[1]{#1}
\csname url@samestyle\endcsname
\providecommand{\newblock}{\relax}
\providecommand{\bibinfo}[2]{#2}
\providecommand{\BIBentrySTDinterwordspacing}{\spaceskip=0pt\relax}
\providecommand{\BIBentryALTinterwordstretchfactor}{4}
\providecommand{\BIBentryALTinterwordspacing}{\spaceskip=\fontdimen2\font plus
\BIBentryALTinterwordstretchfactor\fontdimen3\font minus
  \fontdimen4\font\relax}
\providecommand{\BIBforeignlanguage}[2]{{%
\expandafter\ifx\csname l@#1\endcsname\relax
\typeout{** WARNING: IEEEtran.bst: No hyphenation pattern has been}%
\typeout{** loaded for the language `#1'. Using the pattern for}%
\typeout{** the default language instead.}%
\else
\language=\csname l@#1\endcsname
\fi
#2}}
\providecommand{\BIBdecl}{\relax}
\BIBdecl

\bibitem{Hel2002}
P.~Helgason, ``{Preaspiration in the Nordic Languages. Synchronic and
  Diachronic Aspects},'' Ph.D. dissertation, Department of Linguistics,
  Stockholm University, 2002.

\bibitem{Sil2003}
D.~Silverman, ``On the rarity of pre-aspirated stops,'' \emph{Journal of
  Linguistics}, vol.~39, pp. 575--598, 2003.

\bibitem{Clay2015}
M.~Clayards and T.~Knowles, ``Prominence enhances voiceless-ness and not place
  distinction in {English} voiceless sibilants,'' in \emph{Proceedings of the
  18th International Congress of Phonetic Sciences (ICPhS), Glasgow}, 2015.

\bibitem{dCan2012}
C.~di~Canio, ``The phonetics of fortis and lenis consonants in {Itunyoso}
  {Trique},'' \emph{International Journal of American Linguistics}, vol.~78,
  no.~2, pp. 239--272, 2012.

\bibitem{DF1999}
G.~Docherty and P.~Foulkes, ``Derby and {Newcastle}: instrumental phonetics and
  variationist studies,'' in \emph{Urban Voices: Accent Studies in the British
  Isles}, P.~Foulkes and G.~Docherty, Eds.\hskip 1em plus 0.5em minus
  0.4em\relax Routledge, 1999, pp. 47--71.

\bibitem{GS2010}
O.~Gordeeva and J.~Scobbie, ``Preaspiration as a correlate of word-final voice
  in {Scottish} {English} fricatives,'' in \emph{Turbulent Sounds: an
  Interdisciplinary Guide}, 2010, pp. 167--207.

\bibitem{H2015}
M.~Hejn\'a, ``{Preaspiration in Welsh English: A Case Study of Aberystwyth},''
  Ph.D. dissertation, Department of Linguistics, University of Manchester,
  2015.

\bibitem{HSc2015}
M.~Hejn\'a and J.~Scanlon, ``New laryngeal allophony in {Manchester}
  {English},'' in \emph{Proceedings of the 18th International Congress of
  Phonetic Sciences (ICPhS), Glasgow}, 2015.

\bibitem{JL2003}
J.~J. Jones and C.~Llamas, ``Fricated pre-aspirated /t/ in {Middlesbrough}
  {English}: an acoustic study,'' in \emph{Proceedings of the 15th
  International Congress of Phonetic Sciences (ICPhS), Barcelona}, 2003, pp.
  655--658.

\bibitem{Ket2015}
T.~Kettig, ``{The BAD-LAD Split: a Phonetic Investigation},'' Ph.D.
  dissertation, Department of Theoretical and Applied Linguistics, University
  of Cambridge, 2015.

\bibitem{R1998}
M.~Roos, ``Preaspiration in {Western} {Yugur} monosyllables,''
  \emph{Turgologica}, vol.~32, pp. 28--41, 1998.

\bibitem{S2010}
M.~Stevens, ``How widespread is preaspiration in {Italy}? a preliminary
  acoustic phonetic overview,'' \emph{Lund University Centre for Languages and
  Literature Phonetics Working Papers}, vol.~54, pp. 97--102, 2010.

\bibitem{SH2004}
M.~Stevens and J.~Hajek, ``How pervasive is preaspiration? investigating
  sonorant devoicing in {Sienese} {Italian},'' \emph{Proceedings of the 10th
  Australian International Conference on Speech Science and Technology}, pp.
  334--339, 2004.

\bibitem{W2007}
K.~Watson, ``{The Phonetics and Phonology of Plosive Lenition in {Liverpool}
  {English}},'' Ph.D. dissertation, University of Lancaster, Edge Hill College,
  2007.

\bibitem{Dwyer2000}
A.~Dwyer, ``Consonantalization and obfuscation,'' \emph{Turgologica}, vol.~46,
  pp. 423--432, 2000.

\bibitem{RD2013}
C.~Ringen and W.~A. van Dommelen, ``Quantity and laryngeal contrasts in
  {Norwegian},'' \emph{Journal of Phonetics}, vol.~41, pp. 479--490, 2013.

\bibitem{NSS2013}
C.~Nance and J.~Stuart-Smith, ``Pre-aspiration and post-aspiration in
  {Scottish} {Gaelic} stop consonants,'' \emph{Journal of the International
  Phonetic Association}, vol.~43, no.~2, pp. 129--152, 2013.

\bibitem{H2016a}
M.~Hejn\'a, ``Multiplicity of the acoustic correlates of the fortis-lenis
  contrast: plosives in {Aberystwyth} {English},'' in \emph{Proceeding of
  Interspeech}, 2016, pp. 3147--3151.

\bibitem{RH2008}
P.~Helgason and C.~Ringen, ``Voicing and aspiration in {Swedish} stops,''
  \emph{Journal of Phonetics}, vol.~36, pp. 607--628, 2008.

\bibitem{Hel2004}
P.~Helgason, ``The perception of medial stop contrasts in {Central} {Standard}
  {Swedish}: a pilot study,'' \emph{Proceeding of FONETIK 2004, Stockholm}, pp.
  92--95, 2004.

\bibitem{vDomm1998}
W.~A. van Dommelen, ``Production and perception of preaspiration in
  {Norwegian},'' in \emph{Proceeding of FONETIK 1998, Stockholm}, 1998.

\bibitem{sonderegger2012automatic}
M.~Sonderegger and J.~Keshet, ``Automatic measurement of voice onset time using
  discriminative structured prediction),'' \emph{The Journal of the Acoustical
  Society of America}, vol. 132, no.~6, pp. 3965--3979, 2012.

\bibitem{adi2016automatic}
Y.~Adi, J.~Keshet, E.~Cibelli, E.~Gustafson, C.~Clopper, and M.~Goldrick,
  ``Automatic measurement of vowel duration via structured prediction,''
  \emph{The Journal of the Acoustical Society of America}, vol. 140, no.~6, pp.
  4517--4527, 2016.

\bibitem{adi2016prevoicing}
Y.~Adi, J.~Keshet, O.~Dmitrieva, and M.~Goldrick, ``Automatic measurement of
  voice onset time and prevoicing using recurrent neural networks,'' in
  \emph{Proceeding of the 17th Annual Conference of the International Speech
  Communication Association (Interspeech)}, 2016.

\bibitem{dissen2016formants}
Y.~Dissen and J.~Keshet, ``Formant estimation and tracking using deep
  learning,'' in \emph{Proceeings of the 17th Annual Conference of the
  International Speech Communication Association (Interspeech)}, 2016.

\bibitem{adi2017segmentor}
Y.~Adi, J.~Keshet, E.~Cibelli, and M.~Goldrick, ``Sequence segmentation using
  joint {RNN} and structured prediction models,'' in \emph{Proceeding of the
  42st IEEE International Conference in Acoustic, Speech and Signal Processing
  (ICASSP)}, 2017.

\bibitem{sha2004real}
F.~Sha and L.~K. Saul, ``Real-time pitch determination of one or more voices by
  nonnegative matrix factorization,'' in \emph{Proceeding of NIPS}, 2004, pp.
  1233--1240.

\bibitem{talkin1995robust}
D.~Talkin, ``{A robust algorithm for pitch tracking (RAPT)},'' in \emph{Speech
  coding and synthesis}, W.~Kleijn and K.~Paliwal, Eds.\hskip 1em plus 0.5em
  minus 0.4em\relax New York: Elsevier, 1995, pp. 495--518.

\bibitem{taskar2003max}
B.~Taskar, C.~Guestrin, and D.~Koller, ``{Max-margin Markov networks},'' in
  \emph{Proceeding of NIPS}, 2003.

\bibitem{tsochantaridis2004support}
I.~Tsochantaridis, T.~Hofmann, T.~Joachims, and Y.~Altun, ``{Support vector
  machine learning for interdependent and structured output spaces},'' in
  \emph{Proceeding of the 21st International Conference on Machine Learning
  (ICML)}, 2004.

\bibitem{crammer2006online-short}
K.~Crammer, O.~Dekel, J.~Keshet, S.~Shalev-Shwartz, and Y.~Singer, ``{Online
  passive-aggressive algorithms},'' \emph{Journal of Machine Learning
  Resaerch}, vol.~7, pp. 551--585, 2006.

\end{thebibliography}
